\documentclass[tinyml]{acmart}

\AtBeginDocument{%
  \providecommand\BibTeX{{%
    \normalfont B\kern-0.5em{\scshape i\kern-0.25em b}\kern-0.8em\TeX}}}

\setcopyright{rightsretained}
\copyrightyear{2023}
\acmYear{2023}

\usepackage{amsmath,amsfonts}
\usepackage{xcolor}
\usepackage{pgfplots}
\usepackage{tikz}
\usepackage{adjustbox}
\usepackage{booktabs}
\usepackage[super]{nth}
\usepackage{multirow}
\usepackage{tabularx}

\newcolumntype{P}[1]{>{\centering\arraybackslash}p{#1}}

\begin{document}

\title{FMAS: Fast Multi-Objective SuperNet Architecture Search for Semantic Segmentation}



\author{Zhuoran Xiong}
\email{zhuoran.xiong@mail.mcgill.ca}
\affiliation{%
  \institution{McGill University}
  \streetaddress{address}
  \city{Montreal}
  \country{Canada}
}

\author{Marihan Amein}
\email{marihan.amein@mail.mcgill.ca}
\affiliation{%
  \institution{McGill University}
  \streetaddress{address}
  \city{Montreal}
  \country{Canada}
}

\author{Olivier Therrien}
\email{olivier.therrian@mail.mcgill.ca}
\affiliation{%
  \institution{McGill University}
  \streetaddress{address}
  \city{Montreal}
  \country{Canada}
}

\author{Warren J. Gross}
\email{warren.gross@mcgill.ca}
\affiliation{%
  \institution{McGill University}
  \streetaddress{address}
  \city{Montreal}
  \country{Canada}
}

\author{Brett H. Meyer}
\email{brett.meyer@mcgill.ca}
\affiliation{%
  \institution{McGill University}
  \streetaddress{address}
  \city{Montreal}
  \country{Canada}
}

\renewcommand{\shortauthors}{shortauthors}

\begin{abstract}
We present FMAS, a fast multi-objective neural architecture search framework for semantic segmentation.
FMAS subsamples the structure and pre-trained parameters of DeepLabV3+, without fine-tuning, dramatically reducing training time during search.
To further reduce candidate evaluation time, we use a subset of the validation dataset during the search.
Only the final, Pareto non-dominated, candidates are ultimately fine-tuned using the complete training set. 
We evaluate FMAS by searching for models that effectively trade accuracy and computational cost on the PASCAL VOC 2012 dataset.
FMAS finds competitive designs quickly, e.g., taking just 0.5 GPU days to discover a DeepLabV3+ variant that reduces FLOPs and parameters by 10$\%$ and 20$\%$ respectively, for less than 3\% increased error. We also search on an edge device called GAP8 and use its latency as the metric. FMAS is capable of finding 2.2$\times$ faster network with 7.61\% MIoU loss. 
\end{abstract}



\maketitle

\section{Introduction}

Semantic image segmentation \cite{semanticImageSeg} is one of the fundamental applications in computer vision: 
it helps us understand scenes by identifying the various objects in an image, and their corresponding locations, by predicting an independent class label for each pixel. 
Image segmentation is essential to many applications that run on resource-constrained embedded hardware, such as: 
autonomous driving, medical imaging, and biometric authentication.

Convolutional neural networks (CNN) that achieve state-of-the-art (SOTA) results in image segmentation have sophisticated structures that are generally optimized for accuracy.
They also often require larger feature maps than image classification tasks to be able to produce pixel-wise labels, resulting in a large memory footprint.
In some cases, e.g., autonomous vehicles, image segmentation must be performed in real-time, which makes the development of efficient models for deployment to edge devices critical.

Multi-objective network architecture search has been proposed for the purpose of finding efficient models, but the time required to train candidates is prohibitive. 
DPC~\cite{DPC}, for instance, requires 2,590 GPU days to converge to good designs.
SqueezeNAS~\cite{SqueezeNAS} and TASC~\cite{Template-Based} require 14 and 16 GPU days, respectively.
Fully training a DeepLabV3+ variant with a Modified Xception backbone requires 0.88 GPU days on average, and an additional 0.014 GPU days for evaluation on the validation data set. 
This makes it challenging to search for efficient variants of off-the-shelf networks in a practical amount of time. 
Once-for-all~\cite{Once-For-All} (OFA) solves this problem by essentially training a supernetwork and all its possible subnetworks simultaneously, at an additional computational cost.

To accelerate NAS for image segmentation, we propose Fast Multi-objective Architectural Search (FMAS), a fast multi-objective NAS framework for semantic image segmentation at the edge.
FMAS uses DeepLabV3+ as a supernet to search for computationally-efficient models.
DeepLabV3+~\cite{deeplabv3+} (DL3+) is a SOTA encoder-decoder CNN which employs backbones like Modified Xception~\cite{deeplabv3+} or MobileNetV2~\cite{mobilenetv2} for feature extraction.
FMAS addresses the computational complexity of supernet architectural search in two key ways.
First, candidate networks evaluated by FMAS sub-sample the pre-trained weights of DL3+.
FMAS uses the resulting model performance---\emph{without fine-tuning}---and computational cost, to direct the search toward models with advantageous accuracy-cost trade-offs;
fine-tuning is only performed on final set of Pareto-optimal models. 
Second, FMAS evaluates candidates on a subset of the validation set. 
Neither optimization significantly affects NAS decision making, but together they dramatically reduce search time.

We conducted experiments with PASCAL VOC 2012~\cite{PASCALVOC} to evaluate the effectiveness of our FMAS. 
While reducing search time by 99$\%$ by using weight sharing and a subset of the validation set, our method is capable of finding models with less than a $3\%$ increase in MIoU error.
Moreover, we demonstrate the deployment of image segmentation to TinyML-class~\cite{RAY20221595} systems by targeting the ultra-low-power GAP8~\cite{8445101} platform.
FMAS finds a sub-network model that is 2.2$\times$ faster on GAP8 with a loss in MIoU of 7.61\% compared to the supernet.

\begin{figure*}[t]
  \centering
  \includegraphics[width=0.55\textwidth]{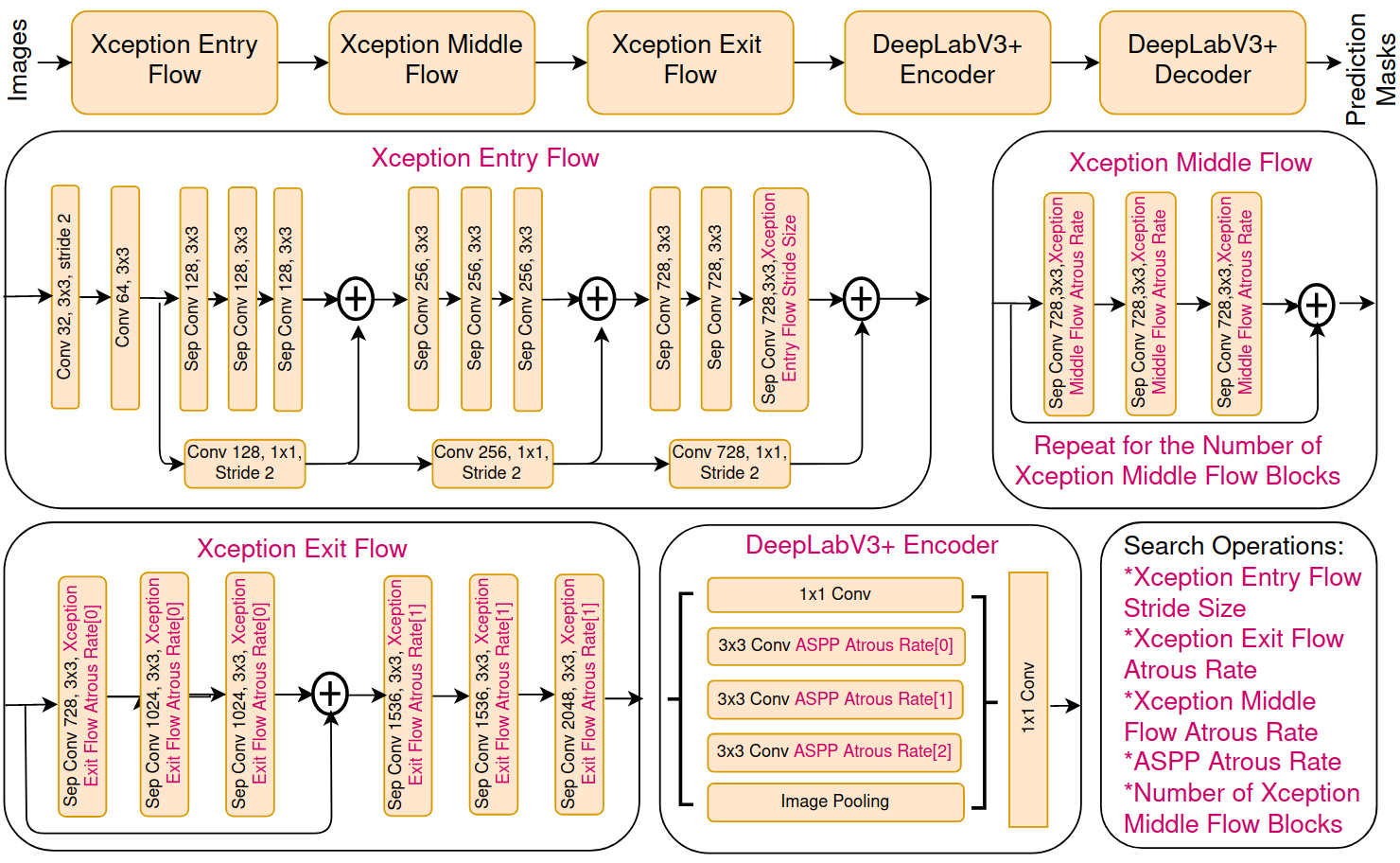}
  \caption{The supernet DeepLabV3+ of FMAS. The base structure is an encoder-decoder preceded with a Modified Xception backbone~\cite{deeplabv3+}. The search space operations and how they are decoded into the supernet structure are highlighted.}
  \label{Computationally-Efficient DeepLabV3+ search space}
\end{figure*}

Specifically, we make the following contributions:
1) We employ NSGA-II~\cite{nsga2}, an elitist genetic algorithm to quickly search for efficient sub-structures in a pre-trained supernet that has been optimized for the targeted task;
2) To accelerate the search, we use a subset of the validation dataset as a proxy; and b) only fine-tune Pareto-optimal models; 
3) We optimize semantic segmentation for edge inference, deploying models on the GAP8 platform. 
To the best of our knowledge, this is the first work to perform NAS for semantic segmentation on ultra-low-power hardware.

\section{Related Work}

To cut search time, multi-objective NAS methods for image segmentation represent the search space in either a \emph{hierarchical}~\cite{Auto-DeepLab},
or a \emph{template-based} way~\cite{DPC, SqueezeNAS, Template-Based}. 
A hierarchical search performs an alternating bi-level optimization, starting by first optimizing the high-level design, then optimizing the internal structure. 
We adopt a template-based search, which tends to converge faster than hierarchical search, as it constrains the high-level structure of the candidates to a pre-defined architecture. 


A number of Reinforcement Learning (RL) NAS approaches for image segmentation have been proposed \cite{Template-Based, fastnas, video}. 
RL-based NAS tends to converge slowly, as it prioritizes long-term reward.
Other approaches~\cite{DPC, Auto-DeepLab, SqueezeNAS} relax the architectural parameters in the search space into a continuous, differentiable form, as gradient descent converges faster. 
It is difficult, however, to represent candidate cost, whether latency or FLOPs, in differentiable form.
Consequently, some approaches~\cite{selfadaptive, EMONAS, nsgaNet} use elitist GAs that consider Pareto-dominance in selection.
We likewise employ NSGA-II.


Several approaches have been proposed to quickly evaluate the accuracy of candidate models. 
Chen et al.~\cite{DPC} designed proxy networks, simplified variants of the candidates that are faster to train and representative of the original accuracy. 
Although this saves some time, they still require 2,590 GPU days.
Alternatively, we use weight sharing, adopting a supernet design space structure, as in~\cite{RONAS, enas, SqueezeNAS}; the candidates use the pre-trained weights of DL3+.
We do not fine-tune candidates during the search at all, dramatically reducing search time without misrepresenting candidate accuracy.

Early stopping, adopted by~\cite{DPC, SqueezeNAS}, assumes networks that train faster are more likely to perform better when fully-trained.
We adopt early stopping, but only during evaluation on the validation set. 
We observe that using a subset of the validation set reduces evaluation time without misrepresenting candidate accuracy.


TinyML~\cite{RAY20221595} refers to running machine learning on edge devices with strict resource constraints.
Prior work~\cite{MLSYS2021_a3c65c29,10.1145/3370748.3406588,9458520} has addressed many different tasks, but none so far have attempted semantic segmentation due to the size of intermediate feature maps and the limited memory of TinyML hardware. 
E.g., U-Net~\cite{unet} requires 61MB of RAM. 
This makes deployment on TinyML hardware like the GAP8, which typically has 8MB of RAM, impossible.
\cite{DBLP:journals/corr/abs-2104-14623} proposed a compact network for semantic segmentation on edge devices.
However, it uses MAC and parameter count, and weight memory, as proxies for cost.
We optimize a model that is too large for TinyML deployment to reduce its memory footprint, and improve inference latency, without substantially degrading accuracy.

\section{FMAS}

We present FMAS, a fast multi-objective NAS framework that uses an elitist genetic algorithm, NSGA-II~\cite{nsga2}, to search for efficient semantic segmentation networks. 
FMAS treats DeepLabV3+, illustrated in Figure~\ref{Computationally-Efficient DeepLabV3+ search space}, 
as a supernet, sub-sampling and reusing its pre-trained parameters.
To further reduce the cost of network evaluation, candidates are evaluated on a subset of the validation dataset.
Only the final, Pareto non-dominated, candidates are fine-tuned using the complete training and validation sets.
We evaluate efficiency by counting parameters and FLOPs, and measuring inference latency on a TinyML platform, the GAP8~\cite{8445101}.

FMAS begins by sampling a population of $M$ model candidates from the design space.
$M-1$ are derived from the structure of DeepLabV3+, and use its pre-trained weights.
The final model is DeepLabV3+ itself, ensuring that the initial population includes a high-accuracy model to accelerate the search.
\emph{No model training occurs at this time.}
The models are then efficiently evaluated, using a fraction of the original validation set.
The parents of the next generation are identified using NSGA-II's selection criteria (which has a strong preference for non-dominated candidates), then crossover and mutation are performed.
Once again, no training is necessary, only accelerated validation, as all network parameters are derived directly from DeepLabV3+.
This process repeats for $G$ generations; the final set of non-dominated models are fine-tuned for $E$ epochs.

\subsection{Subsaming the DeepLabV3+ Architecture}

We constrain the high-level network architecture of the candidates to the structure of 
DeepLabV3+~\cite{deeplabv3+}, using it as a supernet for architecture search.
We experiment with both Modified Xception and MobileNetV2 backbones to perform feature extraction~\cite{deeplabv3+}. 
Then, the DeepLabV3+ encoder-decoder structure classifies each pixel. 
The encoder is based on Atrous Spatial Pyramid Pooling (ASPP)~\cite{deeplab}, which harnesses filters with multiple atrous rates and combines their output feature maps to capture multi-scale features in objects and their context. 
The structure of the decoder is based on bi-linear up-sampling, concatenation, and convolution. 

\subsubsection{DeepLabV3+ with Modified Xception}

Modified Xception 
is based on the structure of the Xception~\cite{xception} network and exhibits improved image segmentation performance. 
Notably, it uses sixteen instead of eight middle flow blocks. 
These computationally-heavy, repeated, structures lend themselves well to parameterized exploration.
We also notice that we can optimize other hyper-parameters that can: improve the accuracy, such as the atrous rates of convolutions; and, reduce the computational cost, such as convolution stride sizes; both without requiring additional trainable weights.

\begin{table}[b]
  \caption{Xception Hyperparameter Design Space}
  \label{The Search Space Opetarions}
  \begin{adjustbox}{center}
  \begin{tabular}{lc}
    \toprule
    Hyperparameter& Possible Values\\
    \midrule
    Xception Entry Flow Stride & 1, 2, 3, 4\\
    Xception Middle Flow Atrous Rate & 1, 2, 3, 4\\
    Xception Exit Flow Atrous Rates & (1, 2), (2, 4)\\
    ASPP Atrous Rates    & (6, 12, 18), (12, 24, 36)\\
    Xception Middle Flow Blocks & $( b_1, b_2, \ldots, b_{16} ), b_i \in \{0, 1\}$\\
  \bottomrule
  \end{tabular}
  \end{adjustbox}
  \end{table}


Table~\ref{The Search Space Opetarions} lists our hyperparameter choices.
We optimize the computational cost of the model by selectively including blocks in the  Xception Middle Flow.
This is the most computationally-expensive section of DeepLabV3+: it requires 65G FLOPs (64\% of DL3+ FLOPs) and 23M parameters (56\% of DL3+ parameters). 
Each of the 16 blocks may be included ($b_i = 1$) or excluded ($b_i = 0$).
To create further opportunity to reduce the computational cost, we optimize the Xception Entry Block Stride Size. 
We also search the atrous rates in the Xception Middle Flow, Xception Exit Flow, and ASPP, as they have the potential to improve segmentation accuracy by widening the receptive field of the images.

Since the middle flow blocks are repeated, the Xception Middle Block Atrous Rates are the same in all selected blocks.
We only search for Xception Entry Block Stride Size in the last convolution stage in the Xception Entry Flow to avoid having the output prediction masks heavily down-sampled by multiple stages of striding, which could result in bad accuracy for semantic segmentation.
The choice sets for the atrous rates in the Xception Exit Block and the ASPP module, listed in Table \ref{The Search Space Opetarions}, are adopted from \cite{deeplabv2}.

Each network is represented with a fixed-length genome consisting of 22 bits, encoding five architectural hyperparameters (Table \ref{The Search Space Opetarions}).
Xception Entry Flow Stride and Xception Middle Flow Atrous Rate both have four design choices, each encoded in two bits.
Each of the Xception Exit Flow Atrous Rates and ASPP Atrous Rates has two sets of design choices, each encoded in one bit. 
Each of the remaining 16 bits indicates the presence (1) or absence (0) of a block of the Xception Middle Flow Blocks.
Any mutation at any bit position results in a legal configuration; likewise, cross-over of any two configurations results in a legal configuration.

\subsubsection{DeepLabV3+ with MobileNetV2}

DeepLabV3+ can also be implemented with MobileNetV2 as its backbone. 
The MobileNetV2~\cite{mobilenetv2} backbone uses fully convolution layers and 19 residual bottleneck layers for its feature extractor.
The 19 residual bottleneck layers are separated into five groups with increasing numbers of output channels, 24, 32, 64, 96, and 160 channels. 
Each group consists of several identical residual bottleneck layers.
Similar to the Modified Xception backbone, we reduce the network by parameterized exploration. 
We search for the dilation rates of depthwise convolutions of six selected residual layers. 
Then, we sampled the convolutional stride of the four selected residual layers, ensuring other non-selected layers are functional, like we do with the Modified Xception backbone.
We also focus our search on the number of repeated identical residual layers for each group. 

Table~\ref{SNAS Architectural Hyperparameters Design Space of MobileNetV2 backbone} lists our hyperparameter choices for the MobileNetV2 backbone. 
We search for the parameters of the residual bottleneck layers in the five groups. 
The value of the number of layers of each group is selected between 1 and the original number of layers of that group. 
We also search for the stride, and the depthwise convolution dilation rate, of the selected residual bottleneck layers.

\begin{table}[b]
  \caption{MobileNetV2 Hyperparameters Design Space}
  \begin{adjustbox}{center}
  \label{SNAS Architectural Hyperparameters Design Space of MobileNetV2 backbone}
  \begin{tabular}{lc}
    \toprule
    Hyperparameter& Possible Values\\
    \midrule
    \nth{2} \& \nth{3} Layer Stride & 2, 3\\
    \nth{14} \& \nth{17} Layer Stride & 1, 2\\
    \nth{12}-\nth{14} Layer Dilation Rate & 1, 2\\
    \nth{15}-\nth{17} Layer Dilation Rate & 1, 2, 3, 4\\
    24-channel Group Layers & $( b_1, b_2), b_i \in \{0, 1\},\sum b_i>0$\\
    32-channel Group Layers & $( b_1, b_2, b_3 ), b_i \in \{0, 1\},\sum b_i>0$\\
    64-channel Group Layers & $( b_1, b_2, b_3, b_4 ), b_i \in \{0, 1\},\sum b_i>0$\\
    96-channel Group Layers & $( b_1, b_2, b_3 ), b_i \in \{0, 1\},\sum b_i>0$\\
    160-channel Group Layers & $( b_1, b_2, b_3 ), b_i \in \{0, 1\},\sum b_i>0$\\
  \bottomrule
  \end{tabular}
  \end{adjustbox}
  \end{table}


\subsection{Evaluating the Accuracy of Candidates}

In order to reduce search time, we employ two strategies targeting candidate training and validation.
First, we adopt the pre-trained weights from DeepLabV3+ and share them with the matching layers of the candidates. 
For the Xception Middle Flow Blocks and Inverted Residual Layers, only the weights for the selected blocks are shared. 
Changing the atrous rates, dilation rates, and stride size does not change the trainable weights in the model.
The intuition of this approach is that if a model is sub-sampled from a supernet, then sharing the network parameters of the supernet and fine-tuning can replace from-scratch training, as demonstrated by~\cite{enas, SqueezeNAS}.

We find that adopting pre-trained weights from DL3+ saves us 0.88 GPU days compared to training a candidate from scratch.
To further reduce evaluation time, we evaluate the candidates on a subset of the validation set. 
We observe that the MIoU error of the models starts to stabilize after being evaluated on 20\% of the validation set. 
Therefore, we use the first 20\% of the validation set. 
After the search is completed, the final Pareto non-dominated models are fine-tuned on the entire training set and re-evaluated on the entire validation dataset.

\section{Experimental Setup}

We conduct two experiments on both Xception and MobileNetV2 on PASCAL VOC to demonstrate how quickly FMAS can find computationally efficient alternatives to DeepLabV3+.
We run our experiments on a Tesla P100-PCIE-16GB GPU.
In the first, we measure the GPU time reduction that results when FMAS is applied to a population of 12 designs on PASCAL VOC 2012.  
We conduct three independent multi-objective searches, for MIoU error and one for each of the computational cost objectives: 
    FLOPs, parameter count, or latency.
In each case, we search a design space with approximately 4M alternatives for Xception and 8M alternatives for MobileNetV2, all derived from DeepLabV3+ and using its pre-trained weights.
For the MobileNetV2 backbone, we evaluate the network's inference latency on the GreenWaves GAP8 SoC using GVSoC~\cite{Bruschi_2021}.
In the second experiment, we fine-tune the final Pareto non-dominated models and re-evaluate their accuracy on the complete dataset after 20 and 25 generations for Xception and MobileNetV2 respectively.
(The appropriate number of generations to use was determined experimentally by quantifying the improvement in Hyperarea difference between Pareto-fronts from one generation to the next, and trading off solution improvement and search time.)
We compare the MIoU result before and after fine-tuning to determine how the MIoU of Pareto non-dominated models is affected by fine-tuning.

In order to fit into the limited memory and supported operations of the GAP8 processor, we (1) scale down the size of the input image to 384$\times$384 pixels, (2) prune the original five branches of the encoder to two branches, and (3) change some of the operations of the original model into GAP8-supported operations, including changing $\mathit{Conv2DTranspose}$ into $\mathit{UpSampling2D}$, and $\mathit{GlobalAveragePooling2D}$ into consecutive $\mathit{AveragePooling2D}s$. 
Together, these changes reduce MIoU by 4.44\% compared with DL3+; pruning encoder branches had the most significant effect on memory footprint. 
The inference on GAP8 and the evaluation on GPU are done in parallel to reduce the search time of the whole process.

\subsection{Candidate Evaluation Metrics} 

FMAS optimizes models to minimize mean intersection over union error (MIoU).
MIoU, adopted by~\cite{deeplabv3+} to evaluate DL3+, is widely used to benchmark the accuracy of semantic segmentation.
In addition to minimizing error, FMAS jointly optimizes models to minimize one of three computational cost metrics.
We quantify the cost of candidate models analytically by measuring either (a) the number of floating-point operations (FLOPs), using keras-flops~\cite{tf-etal}, (b) the number of network parameters, using count\_params, or (c) the latency on the GAP8 processor.

\subsection{PASCAL VOC 2012}

We conduct our experiments with PASCAL VOC 2012~\cite{PASCALVOC}.
PASCAL VOC is a widely used image segmentation dataset with annotated images for 20 object classes, and one background class.
The object classes fall into four categories: person, vehicle, animal, and indoor.
PASCAL VOC contains 1464 training images, and 1449 for validation.
We use the same label image encoding used by DL3+~\cite{deeplabv3+}. 
We use $513 \times 513$ images~\cite{deeplabv3+} for Modified Xception and $384 \times 384$ for MobileNetV2 to fit into the GAP8's memory.

\begin{table*}[t]
    \centering
  \caption{Cost and performance with Modified Xception backbone and derived models}
  \label{NAS Search time}
    \begin{tabularx}{\textwidth}{p{2.2cm}P{1cm}P{1cm}P{1.2cm}P{1.4cm}P{2.3cm}P{0.7cm}P{0.7cm}P{0.7cm}P{1.4cm}P{1.5cm}}\toprule
& \multicolumn{5}{c}{Xception Architecture Parameters} & \multicolumn{3}{c}{Cost}&\multicolumn{2}{c}{MIoU Error (\%)}
\\\cmidrule(lr){2-6}\cmidrule(lr){7-9}\cmidrule(lr){10-11}
           & \small Entry Stride& \small Middle Atrous Rate&\small Exit Atrous Rate&\small ASPP Atrous Rate&\small Middle Blocks&\small GPU Days&FLOPs (G)&\small Params (M)&\small Validation Subset (\%) & \small Fine-tuned + Full Validation \\\midrule
        DeepLabV3+~\cite{deeplabv3+} &2&1&(1,2)&(6,12,18)&1111111111111111&-&101.47&41.26&23.14&22.71\\
        DPC~\cite{DPC}&-&-&-&-&-&2600&99.96&42.70&-&\textbf{19.15}\\
        FMAS-F1&3&1&(1,2)&(6,12,18)&1111111011011111&0.68&\textbf{57.88}&38.00&28.88&27.93\\
        FMAS-F2&2&1&(1,2)&(6,12,18)&1111111011001001&0.52&90.92&33.12&24.95&25.21\\
        FMAS-P1&2&1&(1,2)&(6,12,18)&1111011110010100&\textbf{0.49}&87.41&31.5&27.91&26.64\\
        FMAS-P2&2&1&(1,2)&(6,12,18)&1111111011011111&0.65&101.47&38.00&22.68&22.65\\
        FMAS-FP1&2&1&(1,2)&(6,12,18)&1111111011001101&0.68&94.44&34.75&23.77&24.38\\
        FMAS-FP2&2&1&(1,2)&(6,12,18)&1111011010001100&0.80&83.89&\textbf{29.87}&29.72&29.29\\\bottomrule
    \end{tabularx}
  \end{table*}

\begin{table*}[t]
    \centering
    \caption{Cost and accuracy with MobileNetV2 backbone and derived models}
    \begin{tabularx}{\textwidth}{p{2.3cm}P{1.8cm}P{1.8cm}P{1.6cm}P{0.7cm}P{0.7cm}P{0.7cm}P{1.2cm}P{1.8cm}P{1.8cm}}\toprule
    & \multicolumn{3}{c}{MobileNetV2 Architecture Paremeters} & \multicolumn{4}{c}{Cost}&\multicolumn{2}{c}{MIoU Error (\%)}
\\\cmidrule(lr){2-4}\cmidrule(lr){5-8}\cmidrule(lr){9-10}
        & \small Stride& \small Inverted Layers Dilation Rate &\small Inverted Group Layers & \small GPU Days&FLOPs (G)&\small Params (M)&\small Latency (M Cycles)&\small Validation Subset & \small Fine-tuned + Full Validation \\\midrule
        MobileNetV2~\cite{mobilenetv2} &(2,2,1,1)&(2,2,2,4,4,4)&1111111111&-&9.73&2.14&2189&33.03&\textbf{32.61}\\
        FCN-VGG16~\cite{FCN} &-&-&-&-&243.50&134.49&-&-&37.70\\
        FMAS-L1&(2,2,1,2)&(2,2,1,3,4,2)&1111111111&1.46&7.88&2.14&2085&36.94&36.26\\
        FMAS-L2&(2,3,1,1)&(2,2,2,3,2,2)&1111111111&1.46&\textbf{4.62}&2.14&\textbf{1004}&40.61&40.22\\
    \bottomrule
    \end{tabularx}
    \label{MobileNetV2 Search time}
\end{table*}



\subsection{Final Fine-tuning}

We fine-tune the final Pareto non-dominated networks until convergence, about 6-10 epochs, using a learning rate of 1e-6 to ensure that the re-used weights do not diverge, degrading performance.
We use a batch size of eight, and adopt the remaining hyperparameters in~\cite{deeplabv3+}, with small modifications for stable training: we use a weight decay of 4e-5, and learning rate reduction factor of 0.94.

\section{Results}

\subsection{Reducing Training Time}

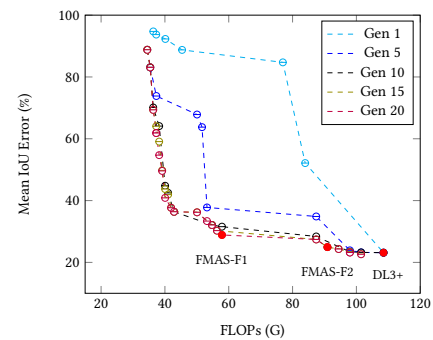
\begin{figure}[b]
\centering
    \begin{tikzpicture}[scale=0.65]
        \begin{axis}[
            enlargelimits=false,
            xlabel={FLOPs (G)},
            ylabel={Mean IoU Error (\%)},
            xmin=15, xmax=120,
            ymin=10, ymax=100,
        ]
        \addplot[
            color=cyan,
            dashed]
        table[x=flops, y=miou]
        {data/pascal/flops/gen1.tex};
        \addplot[
            color=cyan,
            only marks,
            mark=halfcircle,
            mark size=2pt]
        table[x=flops, y=miou]
        {data/pascal/flops/gen1.tex};
        \addplot[
            color=blue,
            dashed]
        table[x=flops, y=miou]
        {data/pascal/flops/gen5.tex};
        \addplot[
            color=blue,
            only marks,
            mark=halfcircle,
            mark size=2pt]
        table[x=flops, y=miou]
        {data/pascal/flops/gen5.tex};
        \addplot[
            color=black, 
            dashed]
        table[x=flops, y=miou]
        {data/pascal/flops/gen10.tex};
        \addplot[
            color=black,
            only marks,
            mark=halfcircle,
            mark size=2pt]
        table[x=flops, y=miou]
        {data/pascal/flops/gen10.tex};
        \addplot[
            color=olive, 
            dashed]
        table[x=flops, y=miou]
        {data/pascal/flops/gen15.tex};
        \addplot[
            color=olive,
            only marks,
            mark=halfcircle,
            mark size=2pt]
        table[x=flops, y=miou]
        {data/pascal/flops/gen15.tex};
        \addplot[
            color=purple, 
            dashed]
        table[x=flops, y=miou]
        {data/pascal/flops/gen20.tex};
        \addplot[
            color=purple,
            only marks,
            mark=halfcircle,
            mark size=2pt]
        table[x=flops, y=miou]
        {data/pascal/flops/gen20.tex};
        \addplot[only marks, red, mark=*, mark size=2pt] coordinates {
        (108.5051695, 23.14418554)
        };
        \node at (axis cs:110, 16){\small DL3+};
        \addplot[only marks, red, mark=*, mark size=2pt] coordinates {
        (57.876716197, 28.877228498458862)
        };
        \node at (axis cs:57.876716197, 21){\small FMAS-F1};
        \addplot[only marks, red, mark=*, mark size=2pt] coordinates {
        (90.924875909, 24.947768449783325)
        };
        \node at (axis cs:90.924875909,17){\small FMAS-F2};
        \legend{{}{Gen 1},{},{}{Gen 5},{},{}{Gen 10},{},{}{Gen 15}, {}, {}{Gen 20}}
        \end{axis}
    \end{tikzpicture}
    \caption{The MIoU-FLOPs Pareto front developed over 20 generations by FMAS on PASCAL VOC 2012.} 
    \label{SNAS-FLOPs on PASCAL}
\end{figure}

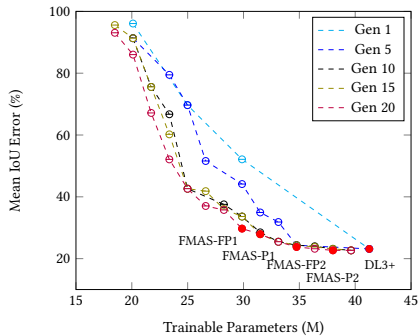
\begin{figure}[t]
\centering
    \begin{tikzpicture}[scale = 0.65]
        \begin{axis}[
            enlargelimits=false,
            xlabel={Trainable Parameters (M)},
            ylabel={Mean IoU Error (\%)},
            xmin=15, xmax=45,
            ymin=10, ymax=100,
        ]
        \addplot[
            color=cyan,
            dashed]
        table[x=params, y=miou]
        {data/pascal/params/gen1.tex};
        \addplot[
            color=cyan,
            only marks,
            mark=halfcircle,
            mark size=2pt]
        table[x=params, y=miou]
        {data/pascal/params/gen1.tex};
        \addplot[
            color=blue,
            dashed]
        table[x=params, y=miou]
        {data/pascal/params/gen5.tex};
        \addplot[
            color=blue,
            only marks,
            mark=halfcircle,
            mark size=2pt]
        table[x=params, y=miou]
        {data/pascal/params/gen5.tex};
        \addplot[
            color=black, 
            dashed]
        table[x=params, y=miou]
        {data/pascal/params/gen10.tex};
        \addplot[
            color=black,
            only marks,
            mark=halfcircle,
            mark size=2pt]
        table[x=params, y=miou]
        {data/pascal/params/gen10.tex};
        \addplot[
            color=olive, 
            dashed]
        table[x=params, y=miou]
        {data/pascal/params/gen15.tex};
        \addplot[
            color=olive,
            only marks,
            mark=halfcircle,
            mark size=2pt]
        table[x=params, y=miou]
        {data/pascal/params/gen15.tex};
        \addplot[
            color=purple, 
            dashed]
        table[x=params, y=miou]
        {data/pascal/params/gen20.tex};
        \addplot[
            color=purple,
            only marks,
            mark=halfcircle,
            mark size=2pt]
        table[x=params, y=miou]
        {data/pascal/params/gen20.tex}; 
        \addplot[only marks, red, mark=*, mark size=2pt] coordinates {
        (41.258213, 23.14418554)
        };
        \node at (axis cs:42.258213, 18){\small DL3+};
        \addplot[only marks, red, mark=*, mark size=2pt] coordinates {
        (31.495733, 27.911359071731567)
        };
        \node at (axis cs:30.495733, 21){\small FMAS-P1};
        \addplot[only marks, red, mark=*, mark size=2pt] coordinates {
        (38.004053, 22.683227062225342)
        };
        \node at (axis cs:38.004053,14){\small FMAS-P2};
        \addplot[only marks, red, mark=*, mark size=2pt] coordinates {
        (29.868653, 29.715323448181152)
        };
        \node at (axis cs:26.868653,26){\small FMAS-FP1};
        \addplot[only marks, red, mark=*, mark size=2pt] coordinates {
        (34.749893, 23.773711919784546)
        };
        \node at (axis cs:34.749893,18){\small FMAS-FP2};
        \legend{{}{Gen 1},{},{}{Gen 5},{},{}{Gen 10},{},{}{Gen 15}, {},{}{Gen 20}}
        \end{axis} 
    \end{tikzpicture}
    \caption{The MIoU-parameters Pareto front developed over the course of 20 generations by FMAS on PASCAL VOC 2012.} 
    \label{SNAS-Params on PASCAL}
\end{figure}

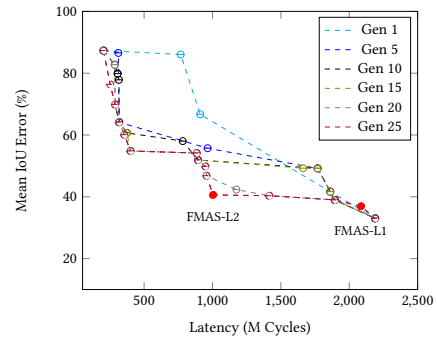
\begin{figure}[t]
\centering
    \begin{tikzpicture}[scale = 0.65]
        \begin{axis}[
            enlargelimits=false,
            xlabel={Latency (M Cycles)},
            ylabel={Mean IoU Error (\%)},
            xmin=50, xmax=2500,
            ymin=10, ymax=100,
        ]
        \addplot[
            color=cyan,
            dashed]
        table[x=latency, y=miou]
        {data/pascal/Latency/gen1.tex};
        \addplot[
            color=cyan,
            only marks,
            mark=halfcircle,
            mark size=2pt]
        table[x=latency, y=miou]
        {data/pascal/Latency/gen1.tex};
        \addplot[
            color=blue,
            dashed]
        table[x=latency, y=miou]
        {data/pascal/Latency/gen5.tex};
        \addplot[
            color=blue,
            only marks,
            mark=halfcircle,
            mark size=2pt]
        table[x=latency, y=miou]
        {data/pascal/Latency/gen5.tex};
        \addplot[
            color=black, 
            dashed]
        table[x=latency, y=miou]
        {data/pascal/Latency/gen10.tex};
        \addplot[
            color=black,
            only marks,
            mark=halfcircle,
            mark size=2pt]
        table[x=latency, y=miou]
        {data/pascal/Latency/gen10.tex};
        \addplot[
            color=olive, 
            dashed]
        table[x=latency, y=miou]
        {data/pascal/Latency/gen15.tex};
        \addplot[
            color=olive,
            only marks,
            mark=halfcircle,
            mark size=2pt]
        table[x=latency, y=miou]
        {data/pascal/Latency/gen15.tex};
        \addplot[
            color=gray, 
            dashed]
        table[x=latency, y=miou]
        {data/pascal/Latency/gen20.tex};
        \addplot[
            color=gray,
            only marks,
            mark=halfcircle,
            mark size=2pt]
        table[x=latency, y=miou]
        {data/pascal/Latency/gen20.tex}; 
        \addplot[
            color=purple, 
            dashed]
        table[x=latency, y=miou]
        {data/pascal/Latency/gen25.tex};
        \addplot[
            color=purple,
            mark=halfcircle,
            mark size=2pt,
            dashed]
        table[x=latency, y=miou]
        {data/pascal/Latency/gen25.tex}; 
        \addplot[only marks, red, mark=*, mark size=2pt] coordinates {
        (2085.645316, 36.94)
        };
        \node at (axis cs:2085.645316, 28.94){\small FMAS-L1};
        \addplot[only marks, red, mark=*, mark size=2pt] coordinates {
        (1004.415432, 40.61)
        };
        \node at (axis cs:1004.415432, 33.61){\small FMAS-L2};
        
        \legend{{}{Gen 1},{},{}{Gen 5},{},{}{Gen 10},{},{}{Gen 15}, {},{}{Gen 20},{},{}{Gen 25}}
    \end{axis}  
    \end{tikzpicture}
    \caption{The MIoU-latency Pareto front developed over the course of 25 generations by FMAS on PASCAL VOC 2012.} 
    \label{SNAS-Latency on PASCAL}
\end{figure}

Figures \ref{SNAS-FLOPs on PASCAL}, \ref{SNAS-Params on PASCAL}, and \ref{SNAS-Latency on PASCAL} plot the MIoU error of the Pareto non-dominated front of the corresponding generation against the FLOPs count, network parameters count, and latency respectively. 
We explore the capacity of FMAS to cut GPU time by evaluating a total of 240 modified Xception variants developed over 20 generations targeting FLOPs, and parameters; and a total of 300 MobileNetV2 variants developed over 25 generations targeting latency.

Table~\ref{NAS Search time} reports the network structures, GPU time consumption, computational cost, accuracy evaluated on a subset of the validation set, and post-fine-tuning accuracy on the entire validation set of selected networks using the Xception backbone.
It presents a selected subset of the final Pareto models produced from Figures~\ref{SNAS-FLOPs on PASCAL} and \ref{SNAS-Params on PASCAL}, and from their combined Pareto fronts, in comparison with DPC~\cite{DPC}.
We observe that FMAS can reproduce the baseline accuracy of DL3+~\cite{deeplabv3+}, achieving MIoU errors of 23\% (e.g., FMAS-FP1), compared to a reported error of 21\% on the validation set.

Similar to Table~\ref{NAS Search time}, Table~\ref{MobileNetV2 Search time} reports results when using the MobileNetV2 backbone and searching for 25 generations.
In addition to FLOPs and parameters, we also report inference latency on the GAP8 for the original model, FCN-VGG16, and selected search results.
Note that while FCN-VGG16 uses only GAP8-supported operations, making it a suitable baseline, it requires more than $8\times$ more RAM than the GAP8 has, and therefore cannot be deployed. 

\subsubsection{Minimizing MIoU Error and FLOPs}

Figure~\ref{SNAS-FLOPs on PASCAL} illustrates the development in the Pareto front with respect to MIoU error and FLOPs over 20 generations.
DL3+ indicates the performance of the baseline with respect to MIoU and FLOPs.
FMAS-F1 and FMAS-F2 are additionally highlighted because of the trade-offs they represent; Table~\ref{The Search Space Opetarions} reports their hyperparameters.
FMAS-F1 cuts the number of FLOPs by 43\% with respect to DL3+, and network parameters by 7.9\%, for a relative increase of 5.2\% in MIoU error; it was discovered in 0.68 GPU days (generation 17).
FMAS-F2 trades off only 2.5\% of the MIoU error of DL3+ for reducing FLOPs by 10\%, and network parameters by 20\%, in 0.52 GPU days (generation 13).

Models in Table~\ref{NAS Search time} required between 0.49 and 0.8 GPU days to be discovered by FMAS, which is negligible compared to the 2,590 GPU days required by DPC~\cite{DPC}.
Although DPC outperforms the MIoU of FMAS-F2 by 6.1\%, FMAS-F2 cuts FLOPs and parameters by 9 and 22\% respectively in only 0.65 GPU days.

We discover FMAS-F2 in 3.5\% the time required to find SqueezeNAS MAC XLarge~\cite{SqueezeNAS}, and 3.1\% of the time required to discover arch0 and arch1 by Nekrasov~\emph{et al.}~\cite{Template-Based}.

\subsubsection{Minimizing MIoU Error and Network Parameters}

Figure~\ref{SNAS-Params on PASCAL} illustrates the development in the Pareto front with respect to MIoU error and network parameters over 20 generations.
In only 0.49 GPU days (generation 12), we find FMAS-P1, which cuts parameters by 24\%, and FLOPs by 14\% for a relative MIoU error increase of 4\%.
FMAS-P2 slightly outperforms the accuracy of DL3+, achieving an MIoU error of 23\%, and also reduces parameters by 7.9\% in 0.65 GPU days (generation 16).
Table~\ref{The Search Space Opetarions} reports their hyperparameters.

We further observe that optimizing for MIoU error and network parameters also produces competitive designs optimized for error, FLOPs, and parameters.
FMAS-FP1 and FP2 were discovered during this search; FMAS-FP1 slightly increases the MIoU error of DL3+ by 1.7\% for a reduction of 6.9\% in the FLOPs count and 16\% in the network parameters count in 0.68 GPU days (generation 12).
FMAS-FP2 reduces FLOPs by 17\% and the parameters by 28\% for an MIoU degradation of 6.6\% in 0.8 GPU days (generation 20).

\subsubsection{Minimizing latency and network parameters}

Figure~\ref{SNAS-Latency on PASCAL} illustrates the development of the Pareto front with respect to MIoU error and latency on the GAP8 over 25 generations. 
In generation 24, we find FMAS-L1 and FMAS-L2.
FMAS-L1 cuts the latency of MobileNetV2 backbone DeepLabV3+ by 4.7\%, 
FMAS-L2 by 54.1\%.
Note that most of this search time is spent compiling for and deploying to the GAP8 for inference latency measurement.
Table~\ref{MobileNetV2 Search time} reports their hyperparameters, FLOPs, parameters, and MIoU error. 
We compare our model with the FCN~\cite{FCN} in terms of the FLOPs and Params. 
While FCN's operations are fully supported by the GAP8, it requires too much memory to be deployed. 

\subsection{Multi-Objective NAS without Fine-Tuning}

In the second experiment, we have selected the final Pareto non-dominated networks that demonstrate the highest accuracy for fined-tuning and re-evaluation on the entire validation set. 
We selected the eight highest-accuracy networks for fine-tuning from Figures~\ref{SNAS-FLOPs on PASCAL},~\ref{SNAS-Params on PASCAL} and the two networks from Figures~\ref{SNAS-Latency on PASCAL}.

After fine-tuning and re-evaluating candidates on the complete validation set, we observed that FMAS either under-estimates or over-estimates the MIoU error by a value that falls in the range between -0.95\% and +1.3\%, as shown in Table~\ref{NAS Search time} and Table~\ref{MobileNetV2 Search time}. 
Such minor changes in accuracy show how FMAS can reliably and quickly evaluate the accuracy of networks sub-sampled from a supernet, without having to fine-tune candidates during the search.

FMAS can bias search toward designs that perform better without training, but it has proven effective in finding efficient designs (e.g. FMAS-F2) in a competitive search time compared to existing multi-objective NAS methods (e.g. DPC and SqueezeNAS). 
Therefore, discarding designs that could potentially perform better is still a worthwhile trade-off given the advantage in search time reduction.

\subsection{Discussion}

Considering the 240 network evaluations and eight fine-tuned networks for each search, FMAS ran each complete search in three GPU days, instead of 220, resulting in an aggregate GPU time reduction of 99\%.
We attribute most of this reduction to re-using the pre-trained weights of DL3+, which eliminates 0.88 GPU days per candidate, saving 215 days.
Evaluating a candidate on the first 20\% of the validation set cuts its evaluation time by 0.014 GPU days. 
For a complete search, this cuts evaluation time by three GPU days.
Fine-tuning each of the final Pareto non-dominated candidates requires approximately 0.14 GPU days.


Several patterns emerge in the hyperparameters selected by the Pareto non-dominated models in our experiments.
Analyzing the Xception Middle Flow Blocks of the final models optimized for FLOPs and MIoU (Figure \ref{SNAS-FLOPs on PASCAL}), blocks 1, 9 and 13 have at least a 90\% chance of being active, whereas block 10 was always absent.
Increasing the Xception Entry Block Stride Size reduces FLOPs, as well as accuracy, compared to other hyper-parameter choices.
The structures of FMAS-F1 and FMAS-P2 are identical except for stride; FMAS-F1 cuts FLOPs by 44\% compared to FMAS-P2, for a relative increase in the MIoU error of 5.2\%. 
The networks in Table~\ref{NAS Search time} adopt the smaller atrous rates for Xception Middle Flow, Xception Exit Flow, and ASPP. 
We observe that the highlighted networks using the MobileNetV2 backbone, FMAS-L1 and FMAS-L2, use all inverted residual blocks. 
The dilation rates of the later layers, e.g. $\nth{15}$ and $\nth{17}$ layer, are more likely to be smaller to reduce inference latency. 

\section{Conclusion}

We present FMAS, a multi-objective NAS framework that significantly reduces the search time for finding efficient semantic segmentation networks. 
FMAS uses NSGA-II to sub-sample candidates from a DeepLabV3+ supernet and improve their performance over the course of generations through crossover and mutation.

We reduce GPU time by re-using the pretrained weights of DL3+ for candidates, and evaluating accuracy on a subset of the validation set. 
We fine-tuned only the final Pareto non-dominated models.
This saved 0.88 GPU days of training time per candidate, and an additional 0.014 GPU days, respectively.
Adding the time required for fine-tuning (0.14 GPU days each), FMAS saved 99\% of the 220 GPU days required to run the entire search.
We also demonstrate that no fine-tuning is required during search. 
We found that the accuracy of the discovered networks before and after fine-tuning differs from -0.95\% to +1.3\%, an inconsequential difference resulting in substantial time savings.
Besides, we prove that our method can be generalized to different networks and different performance metrics by applying FMAS to DeepLabv3+ with two backbones and three metrics. 
The search results of two backbones show that FMAS is capable of finding latency-efficient models.

\section{Acknowledgements}

This research was made possible by the support of: the Natural Sciences and Engineering Research Council of Canada (NSERC), though grant number CRDPJ 531142-18; and, Synopsys Inc.
\clearpage
\bibliographystyle{ACM-Reference-Format}
\bibliography{sample-tinyml}

\end{document}